\documentclass{Interspeech2024}
\usepackage{kotex}
\usepackage{arydshln}
\usepackage{multirow}
\usepackage{makecell}
\usepackage{todonotes}
\usepackage{tipa}
\usepackage{hyperref}
\usepackage{footnote} % for nice tables
\interspeechcameraready

\title{Improving Domain-Specific ASR with LLM-Generated Contextual Descriptions}

\name[affiliation={\dagger}]{Jiwon}{Suh}
\name[affiliation={\dagger}]{Injae}{Na}
\name[affiliation={}]{Woohwan}{Jung}
\address{
  Department of Applied Artificial Intelligence, Hanyang University, Republic of Korea}
\email{\{jwsuh0205, suhoij47, whjung\}@hanyang.ac.kr}

\usepackage[T1]{fontenc}

\newcommand{\minisection}[1]{%
\vspace{0.04in}
    \noindent \textbf{#1}.\xspace%
}

\begin{document}

\maketitle

\begingroup
\renewcommand\thefootnote{†}
\footnotetext{\scriptsize Major in Bio Artificial Intelligence}
\endgroup

\begin{abstract}
End-to-end automatic speech recognition (E2E ASR) systems have significantly improved speech recognition through training on extensive datasets. Despite these advancements, they still struggle to accurately recognize domain specific words, such as proper nouns and technical terminologies. To address this problem, we propose a method to utilize the state-of-the-art Whisper without modifying its architecture, preserving its generalization performance while enabling it to leverage descriptions effectively. Moreover, we propose two additional training techniques to improve the domain specific ASR: decoder fine-tuning, and context perturbation. We also propose a method to use a Large Language Model (LLM) to generate descriptions with simple metadata, when descriptions are unavailable. Our experiments demonstrate that proposed methods notably enhance domain-specific ASR accuracy on real-life datasets, with LLM-generated descriptions outperforming human-crafted ones in effectiveness.        
\end{abstract}

\section{Introduction}

Recent advancements in end-to-end (E2E) automatic speech recognition (ASR) systems, such as Wav2Vec 2.0 \cite{wav2vec} and Whisper \cite{whisper}, have significantly improved the capabilities of speech recognition through extensive training on large datasets.
However, these systems often encounter difficulties in accurately identifying domain specific terms, such as proper nouns and technical jargon.
Consider a scenario where an ASR system is tasked with transcribing a lecture on mycology. 
The pronunciation of `morel' ([m\textopeno\textlengthmark\textprimstress rel]
) and `moral' ([\textprimstress m\textopeno r\textschwa l]) can be very similar, especially in rapid speech, although these words have different meanings.
Since `moral' appears much more frequently in general speech than `morel', general-domain ASRs may misinterpret `morel' as `moral'.

Contextual biasing \cite{aleksic2015bringing} is widely used in previous studies \cite{le2021contextualized,huang2023contextualized,CB-Conformer,sun2023can} to improve the performance of speech recognition for domain-specific words. 
In addition to the audio input, this method provides a biasing list that consists of words that do not appear frequently throughout the dataset or with a high error rate. 
However, a notable limitation arises from the reliance on a single biasing list tailored for each specific dataset, such as Mathematics, Finance, and Chemistry.
Creating a comprehensive list that covers all potential domain-specific terms is inherently challenging. Thus, it often fails to recognize highly specific words in the current audio input (e.g., Creating a comprehensive list that covers all product names across various companies for earnings calls.).

To alleviate the problem, recent works \cite{PromptASR,chang2023context,promptformer,deep-LLM-fusion} introduce methods to incorporate more specific contextual information.
Some of the works \cite{chang2023context,promptformer} utilize the preceding utterance as the contextual information. 
However, this approach can suffer from the error propagation problem: an incorrect recognition in an utterance causes incorrect recognition in the next utterances.
In addition, they require an additional text encoder which should be trained on the domain specific dataset, demanding a substantial amount of training data that is frequently unavailable in domain-specific ASR.
The most relevant work to ours is \cite{deep-LLM-fusion}, which utilizes human-written descriptions.
However, it also employs an additional model LLaMa~\cite{Llama} to utilize the textual description.
Since LLaMa (7B$\sim$65B parameters) is much larger than typical ASR models like Whisper (0.04B $\sim$ 1.6B parameters), it introduces a significant computational overhead.
Moreover, the human-written descriptions in real-world datasets often lack details and even are unavailable in many cases.

In response to these limitations, primarily due to the integration of additional models with pretrained ASR models, 
We utilize the existing Whisper \cite{whisper} framework without any additional module. 
Since Whisper is already trained for ASR tasks, our approach minimizes the requirement for domain-specific training data.
We also introduce several techniques to efficiently fine-tune Whisper for domain-specific ASR with small domain specific data.
Moreover, for the cases without human-created descriptions, we propose a method to generate a description for each audio by using a Large Language Model (LLM).
The cost of utilizing an LLM is relatively small by generating a single description for each speech rather than for each utterance.
Notably, we empirically demonstrate that the LLM-generated description can surpass human-created ones in improving the accuracy of domain-specific ASRs, thanks to the detailed explanations of LLMs.

\section{Method}
In this section, we propose a method\footnote{\scriptsize https://github.com/nickjw0205/Improving-ASR-with-LLM-Description} to improve domain-specific ASR by incorporating textual descriptions.  
\begin{figure*}[tb]
    \centering
    \resizebox{0.90\textwidth}{!}{
        \includegraphics{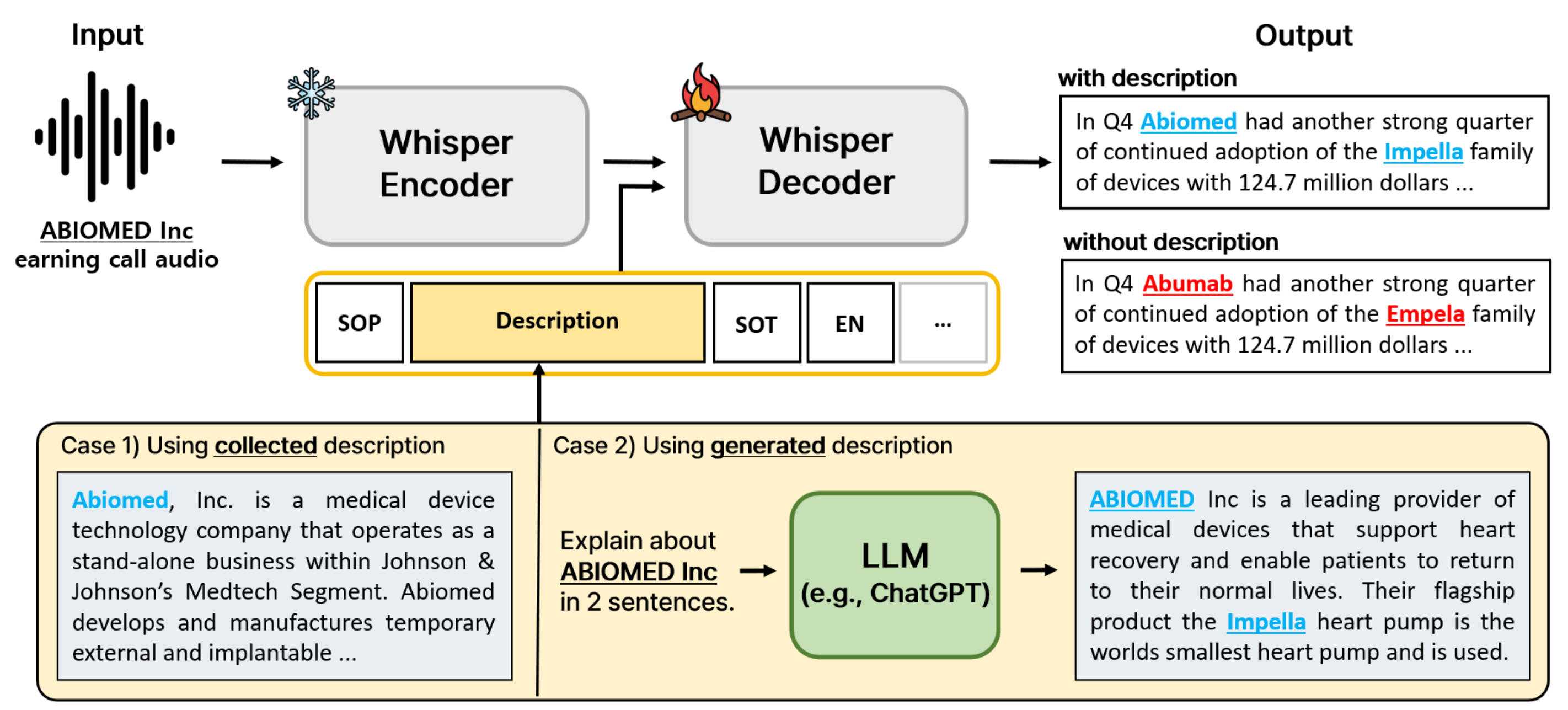}
    }
    \caption{An overview of our method}
    % The utilization of descriptions is divided into two cases: one where descriptions are manually collected (Case 1), and the other where descriptions are generated through LLM due to the inability to collect them manually (Case 2).
    \label{fig:overview_training_resume}
\end{figure*}

\subsection{Overview}
Previous works \cite{PromptASR,chang2023context,promptformer,deep-LLM-fusion} have explored improving ASR models with textual context typically by combining an acoustic model and a language model.
For instance, a recent work \cite{deep-LLM-fusion} leverages  HuBERT \cite{Hubert} as the audio encoder and LLaMa \cite{Llama} as the language model.
Despite the effective utilization of textual information, this approach requires extensive domain-specific data for model training.
To address this challenge, we decide to utilize descriptions without modifying the architecture of an existing ASR model. 

Conventional ASR models usually do not have a text encoder since they primarily focus on converting speech to text.
Thus, we cannot use such a model without additional modules.
However, Whisper \cite{whisper}, a state-of-the-art ASR model, intrinsically supports textual input.
Whisper is built on Transformer \cite{Transformer} architecture with an encoder and a decoder.
The encoder processes audio inputs while the decoder generates transcripts. 
Given that the decoder is based on a Transformer, it can accept a prompt prior to transcript generation. 
Thus, we input textual descriptions as prompts. 

Figure~\ref{fig:overview_training_resume} shows an overview of our method which utilizes a pretrained Whisper model without modifying its original architecture. 
The audio input is processed by the encoder, consistent with the original Whisper.
Note that the encoder remains frozen during fine-tuning, a topic further explored in Section~\ref{sec:training}.
We give textual descriptions as prompts to the decoder.
Whisper has two special tokens \texttt{<SOP>} (Start of previous) and \texttt{<SOT>} (Start of transcript) which is to take the transcription of the preceding 30 seconds of audio segment as contextual information.  
We adapt these tokens to input our textual descriptions in the format
 \texttt{<SOP>} \texttt{description} \texttt{<SOT>}.
This strategy enables the model to utilize descriptions without an additional text encoder.

\subsection{Training the Proposed Model}
\label{sec:training}

Our model can be fine-tuned in the same way as the original Whisper was trained since it uses the architecture of Whisper without any modification. 
However, to improve the performance of the domain-specific ASR tasks with textual description, we need to consider its unique characteristics.
First, domain-specific datasets are generally much smaller than general-domain datasets.
Second, not every utterance directly relates to the provided textual description. %For instance, 
We propose two methods to address the issues: decoder fine-tuning and context perturbation.

\minisection{Decoder Fine-tuning} Domain-specific ASR models suffer from the scarcity of training data with speech-text pairs. 
To avoid overfitting and catastrophic forgetting, it is necessary to minimize the number of parameters to be trained. 
The decoder of our model process descriptions as additional inputs, a feature not present during the pretraining phase of Whisper, mandating fine-tuning. 
Conversely, fine-tuning the encoder is not essential since it solely processes audio, just as during pretraining. 
The pretrained Whisper encoder has an excellent capability in featuring audio information even without fine-tuning.
According to a recent study \cite{yang2023investigating}, Whisper with a frozen encoder exhibits superior performance in various tasks, including intent classification, keyword spotting, and ASR, compared to the model where the encoder has been fine-tuned.
It implies that retraining the encoder may require a significant amount of data and risk losing its generalization performance. 
Therefore, we opt to freeze the encoder and fine-tune the decoder to effectively leverage the textual description. 

\minisection{Context Perturbation} 
Although textual descriptions aid in recognizing domain specific words, not all utterances contain domain specific words.
For example, situations like greetings or humorous remarks for ice-breaking may not include domain-specific words.
In these instances, ASR models should be able to ignore the provided description.
Inspired by \cite{PromptASR}, 
we occasionally input a randomly sampled description instead of the actual one with a probability of 5\% in our experiments. 
This approach trains the model to selectively utilize the description.

\subsection{Generating Descriptions using LLM}
\label{sec:gen_desc}

Utilizing textual descriptions can significantly improve ASR performance. 
However, there are numerous situations where textual descriptions may not be available, such as personal recordings and live streams. 
Moreover, even when textual descriptions are available, they often do not include necessary detail. 
For example, consider the task of transcribing a university lecture through ASR. 
While the syllabus may provide an overview of the course, it typically lacks specific details about individual lecture topics. 
To overcome this challenge, we propose generating detailed descriptions through LLM.

We generate descriptions by using LLM with simple metadata about audio files.
To validate the effectiveness of LLM-generated descriptions with very simple metadata, 
we use two datasets in our experiments: Earnings Call \cite{Earningscall}, which contains discussions about financial results of particular public companies, and OCW (MIT OpenCourseWare), which includes video lectures.
For the Earnings Call dataset, the description of conference call is created by using the company name with a prompt “Explain about {[\texttt{company name}]} in 2 sentences.”.
Due to the weakness of Whisper with processing long descriptions \cite{zhuo2023lyricwhiz}, we limit the length of the description with the instruction  "in 2 sentences".
For the OCW dataset, we use the prompt “Today’s lecture title is {[\texttt{lecture title}]}. Please explain the academic field and content in 2 lines.”.
This technique, relying solely on simple metadata (lecture titles and company names), enables the extraction of essence-capturing descriptions, drawing on the extensive knowledge of LLM.
It is worth noting that the cost of using LLM is not high.
For example, the OCW dataset has 508 audio segments per lecture on average, indicating that the execution of ASR model is 508 times more frequent than that of LLM in the OCW dataset.

\section{Experiments}
\subsection{Datasets}
To evaluate the domain-specific ASR models with descriptions, we use the following two datasets.

\minisection{Earnings Call \cite{Earningscall}}
It consists of quarterly earning conference calls from S\&P 500 companies in 2017. 
The Earnings Call dataset contains various domain-specific information, such as company and product names. 
After removing duplicate calls from the same companies, 
this dataset consists of 169 conference calls.
We split the dataset into 113, 28, and 28 calls for training, validation, and testing, respectively.
The total lengths of the audio in train/valid/test sets are approximately 40/10/10 hours, respectively.
The audio quality of Earnings Call is much lower than professionally recorded datasets due to factors like multiple speakers and fluctuating recording environments.

\minisection{OCW\footnote{\scriptsize https://github.com/nickjw0205/Improving-ASR-with-LLM-Description} }
We collected a new dataset to validate the effectiveness of our model with academic words.
This dataset is collected from a MIT OpenCourseWare website\footnote{\scriptsize https://ocw.mit.edu/, CC BY-NC-SA 4.0 license}
and includes the videos, English transcripts, and titles of 65 lectures on various academic fields such as linear algebra, biological chemistry, and cryptocurrency.
We split the dataset into train (44 lectures), valid (12 lectures), and test (9 lectures) sets to make their lengths are approximately 40 hours, 10 hours, and 10 hours, respectively.
We excluded all words from the transcript that were not actual utterances but were added to aid understanding. 
Examples of such words include [No audio], [Not AUDIBLE], [Laugh], [DOOR CLOSES], etc. 
Additionally, we performed preprocessing to ensure proper alignment between the audio and transcript by removing the speaker identifiers, such as `Speaker:', used to denote the speaker.

\subsection{Experimental Setup}
We utilize the Whisper (base.en), a Transformer encoder-decoder model with 74 million parameters.
Notably, our method fine-tunes only the decoder which has 52 million parameters.
To ensure compatibility with Whisper, all audio files are resampled to 16kHz and transcriptions are normalized by converting to lowercase and removing punctuations.
To train Whisper, we use the HuggingFace Transformers library\footnote{\scriptsize https://github.com/huggingface/transformers}.
Following the Whisper's training, we use AdamW \cite{adamW} optimizer with a linear learning rate decay scheduler.
The training process was conducted with a batch size of 32, initiating learning rate decay following a 100-step warm-up period. We explored initial learning rates of 1e-4, 1e-5, and 1e-6, ultimately selecting 1e-4 for Earnings Call and 1e-6 for OCW based on performance. 
Training durations are set at 10 epochs for Earnings Call and 15 epochs for OCW.
We use the cross entropy loss following Whisper.
Other settings follow the default configuration of the HuggingFace trainer. 
All models are trained on an NVIDIA RTX 4090 GPU. 
Unless otherwise specified, we use the descriptions generated by GPT-3.5 Turbo\footnote{\scriptsize https://openai.com/product} as proposed in Section~\ref{sec:gen_desc}.
The evaluation metric is the Word Error Rate (WER).
Note that the training time was about 6 hours for Earnings Call, and 8 hours for OCW.

\begin{table}[tb]
\setlength{\abovecaptionskip}{2pt}
\setlength{\belowcaptionskip}{1pt}
\caption{Main results}
\renewcommand{\arraystretch}{1.1}
\resizebox{\columnwidth}{!}{%
\begin{tabular}{lcccc}
\toprule
\multicolumn{1}{c}{} & \multicolumn{2}{c}{Earnings Call} & \multicolumn{2}{c}{OCW} \\ 
Models & 20 h & 40 h & 20 h & 40 h \\
\midrule 
 Whisper (Frozen) & 16.39\% & 16.39\% & 11.98\% & 11.98\% \\ 

+ Full Fine-tuning & 17.38\% & 16.64\%& 10.41\% & 9.94\%   \\

+ Description & 20.63\% & 17.70\% & 9.81\% & 9.72\%   \\

+ Decoder Fine-tuning & 16.61\% & 15.70\% & \textbf{9.79\%} & \textbf{9.67\%}   \\ 
\
+ Context Perturbation& \textbf{16.24\%} & \textbf{15.15}\% & \textbf{9.79\%} & 9.68\%   \\
\bottomrule
\end{tabular}%
}
\label{tab:main-result}
\end{table}

\begin{table*}[tb]
\setlength{\abovecaptionskip}{2pt}
\setlength{\belowcaptionskip}{1pt}
\centering
\caption{Examples of outputs with and without descriptions.}
\begin{tabular}{l p{0.4\textwidth} p{0.4\textwidth}}
\toprule
 & \multicolumn{1}{c}{Edwards Lifesciences (Earnings Call)} & \multicolumn{1}{c}{Assembly Language \& Computer Architecture (OCW)} \\ 
\midrule
Description & Edwards Lifesciences is a global medical technology company that specializes in the development and manufacturing of heart valves ... improve the lives of patients with \textbf{cardiovascular} diseases & Assembly language is a low-level \textbf{programming language} that communicates directly with the hardware of a computer system. Computer architecture is the study of how computers ... \\ 
\midrule
\midrule
Reference & New treatments and therapies for \textbf{\color{blue}{cardiovascular}} care and heart muscle recovery & It's just a bunch of \textbf{\color{blue}{bytes}} \\ 
\midrule
\multirow{1}{*}{\parbox{2cm} {Output \\ w/ description}} & New treatments and therapies for \textbf{\color{blue}{cardiovascular}} care and heart muscle recovery & It's just a bunch of \textbf{\color{blue}{bytes}} \\ 
\midrule
\multirow{1}{*}{\parbox{2cm} {Output \\ w/o description}} & New treatments and therapies for \textbf{\color{red}{cardiovasular}} care and heart missile recovery & It's just a bunch of \textbf{\color{red}{bites}} \\ 
\bottomrule
\end{tabular}
\label{tab:qualitative-analysis}
\end{table*}

\subsection{Main Results}  
To demonstrate the effectiveness of our proposed method we show the word error rates of ASR with different train data sizes (20 and 40 hours) in Table~\ref{tab:main-result}.
We systematically investigate the effectiveness of the proposed components by incrementally introducing the following elements to the frozen Whisper: fine-tuning, utilization of descriptions, decoder fine-tuning (freezing encoder), and context perturbation.
Applying all proposed techniques result in the best ASR model performance, demonstrating the effectiveness of our methods.

The experimental results on Earnings Call (20 hours) are quite interesting. 
When the Whisper model was fully fine-tuned with textual descriptions, there was a significant increase in WER by more than 4 percentage points, indicating a substantial performance decline. 
This can be interpreted as a result of overfitting and catastrophic forgetting, which occurs when training on a small amount of data for domain-specific ASR tasks, as we previously mentioned. 
However, even in such cases, our method, employing decoder fine-tuning and context perturbation, demonstrated its ability to overcome these issues. 
This suggests that it is essential to perform decoder fine-tuning along with context perturbation to achieve robust results with a small amount of data and in conditions of low audio quality. 
While the results from the OCW dataset are not as dramatic as those from the Earnings Call (20h), they still confirm the stable performance improvements by our methods.

\begin{table}[tb]
\setlength{\abovecaptionskip}{2pt}
\setlength{\belowcaptionskip}{1pt}
\centering
\caption{Examples of collected and generated descriptions.}
\renewcommand{\arraystretch}{1.1}
\begin{tabular}{c p{0.73\columnwidth}}
\toprule

 & \multicolumn{1}{c}{Description} \\
\midrule
\multirow{1}{*}{\parbox{1.3cm}{\centering Collected}}  & This course is an introduction to mathematical modeling of computational problems, as well as common algorithms, algorithmic paradigms, and data structures used to solve these problems. It emphasizes the relationship between algorithms and programming, and introduces basic performance measures and analysis techniques for these problems. \\ 
\midrule
\multirow{2}{*}{\parbox{1.3cm}{\centering LLM \\ Generated}} & \textbf{\color{blue}{hashing}} is a fundamental concept in computer science and cryptography that involves creating a unique, fixed-length representation of data using a mathematical algorithm. It is commonly used in data retrieval, password security, and ensuring data integrity. \\ 

\bottomrule
\end{tabular}
\label{Tab:LLM-Description}
\end{table}

\subsection{Qualitative Analysis}
Table~\ref{tab:qualitative-analysis} shows examples of transcription, both with and without using descriptions.
The first example provides a description about Edward Lifesciences, a medical technology company, and transcription results for a sentence from an Edward Lifesciences earnings call. 
The description incorporates domain-specific terminology such as cardiovascular and includes comprehensive information about the company.
By using this description, our method accurately transcribes the medical term `cardiovascular'. 
However, the conventional speech recognition method struggles to accurately transcribe the domain-specific word `cardiovascular'. 
In the OCW dataset, the description offers insights into "Assembly Language and Computer Architecture," facilitating the accurate identification of computer-related terms such as byte' based on this contextual information. Conversely, without utilizing the description, the term `byte' is misrecognized as its homophone `bite', because there is no computer-related context information during the transcription process. These indicate that descriptions provide not only domain-specific terms but also important contextual information to the ASR model. Therefore, incorporating descriptions into the ASR model can enhance the recognition accuracy of domain-specific words.

\subsection{Comparison of LLM-Generated and Collected Descriptions}
\minisection{Collection of human-written descriptions}
To evaluate the effectiveness of the LLM-generated descriptions, we also collect human-written descriptions for our dataset as a comparison.
We collect descriptions from Wikipedia for company information in the Earnings Call dataset, which are often overly detailed because they aim to provide comprehensive explanations. For the OCW dataset, we collect descriptions of the respective courses that provide a broad course overview, rather than specifics of individual lectures, reflecting their purpose of providing a general course summary. 

\minisection{Description Quality}
Table~\ref{Tab:LLM-Description} presents a LLM-generated description and using the lecture title ``Hashing" and a human-written one collected from the course description of ``Introduction to Algorithms".
The collected description offers an overview of the course, outlining the content to be learned at a comprehensive level.
In contrast, the description generated by LLM offers detailed information about the specific lecture `Hashing'.
This indicates that LLM can generate more specific and relevant descriptions than collected descriptions in real world applications.

\begin{table}[tb]
    \centering
    \caption{Results on generated and collected descriptions.}
    \vspace{-0.2cm}
    \begin{tabular}{lccc}
    \toprule
    Description & Earnings Call & OCW \\
    \midrule
     Collected & 15.33\% & 10.30\% \\
     LLM Generated & \textbf{15.15}\% & \textbf{9.68}\% \\
    \bottomrule
    \end{tabular}
    \vspace{-0.2cm}
    \label{tab:LLM-vs-Base}
\end{table}

\minisection{ASR Performance}
In Table~\ref{tab:LLM-vs-Base}, we compare ASR performances with collected descriptions and LLM-generated descriptions. 
For the Earnings Call, transcription results using LLM-generated descriptions slightly outperform those using collected descriptions by 0.18\%p.
The LLM-generated descriptions contribute to this improvement by omitting unnecessary details and adopting more concise and central information.
In the OCW, transcriptions utilizing LLM-generated descriptions demonstrated a lower WER compared to those with collected descriptions by 0.62\%p.
This difference is due to the clear disparity in content between the LLM-generated description and the collected one, as detailed in Table 3.
Collected descriptions provide only general information about the courses, failing to capture the specific details essential for each lecture audio.
However, the LLM-generated description is tailored to the lecture title ``hashing", providing detailed information relevant to the specific lecture. 
\section{Conclusion}

We propose a domain-specific speech recognition model which leverages descriptions of speeches.
To address the data scarcity problem in domain-specific ASR, we propose a method to utilize the state-of-the-art Whisper without modifying its architecture.
Moreover, we propose decoder fine-tuning, and context perturbation to preserve the Whisper's generalization performance while utilizing descriptions. 
The empirical studies show the effectiveness of our methods with two real datasets.
Furthermore, we find that LLM-generated descriptions can  outperform human-written ones in domain-specific ASR with description. 

\clearpage

\bibliographystyle{ieeetr}
\bibliography{ref}
\end{document}